\documentclass{article}

\usepackage{times}
\usepackage{graphicx} % more modern
\usepackage{subfigure} 

\usepackage{natbib}

\usepackage{algorithm}
\usepackage{algorithmic}

\usepackage{hyperref}

\usepackage[accepted]{icml2017}

\usepackage{amssymb,amsmath}
\usepackage{graphicx}
\usepackage{algorithm,algorithmic}
\newcommand{\expB}[1]{\exp\left\{#1\right\}}
\icmltitlerunning{Variational Inference for Sparse and Undirected Models}

\begin{document} 

\twocolumn[
\icmltitle{Variational Inference for Sparse and Undirected Models}



% you can specify symbols, otherwise they are numbered in order

\begin{icmlauthorlist}
\icmlauthor{John Ingraham}{ha}
\icmlauthor{Debora Marks}{ha}
\end{icmlauthorlist}

\icmlaffiliation{ha}{Harvard Medical School, Boston, Massachusetts}
\icmlcorrespondingauthor{John Ingraham}{ingraham@fas.harvard.edu}
\icmlcorrespondingauthor{Debora Marks}{debbie@hms.harvard.edu}

\icmlkeywords{boring formatting information, machine learning, ICML}

\vskip 0.3in
]



\printAffiliationsAndNotice{}  % leave blank if no need to mention equal contribution
%\footnotetext{hi}

\begin{abstract} 
Undirected graphical models are applied in genomics, protein structure prediction, and neuroscience to identify sparse interactions that underlie discrete data. Although Bayesian methods for inference would be favorable in these contexts, they are rarely used because they require doubly intractable Monte Carlo sampling. Here, we develop a framework for scalable Bayesian inference of discrete undirected models based on two new methods. The first is Persistent VI, an algorithm for variational inference of discrete undirected models that avoids doubly intractable MCMC and approximations of the partition function. The second is Fadeout, a reparameterization approach for variational inference under sparsity-inducing priors that captures \emph{a posteriori} correlations between parameters and hyperparameters with noncentered parameterizations. We find that, together, these methods for variational inference substantially improve learning of sparse undirected graphical models in simulated and real problems from physics and biology.
\end{abstract} 
%%%%%%%%%%%%%%%%%%%%%%%%%%% INTRO %%%%%%%%%%%%%%%%%%%%%%%%%%%
\section{Introduction}
Hierarchical priors that favor sparsity have been a central development in modern statistics and machine learning, and find widespread use for variable selection in biology, engineering, and economics. Among the most widely used and successful approaches for inference of sparse models has been $L_1$ regularization, which, after introduction in the context of linear models with the LASSO \cite{tibshirani1996regression}, has become the standard tool for both directed and undirected models alike \cite{murphy2012machine}.  

Despite its success, however, $L_1$ is a pragmatic compromise. As the closest convex approximation of the idealized $L_0$ norm, $L_1$ regularization cannot model the hypothesis of sparsity as well as some Bayesian alternatives  \cite{tipping2001sparse}. %Bayesian sparsity-promoting priors can yield considerably better results \cite{tipping2001sparse}, but only when Bayesian inference is tractable. 
Two Bayesian approaches stand out as more accurate models of sparsity than $L_1$. The first, the spike and slab \cite{mitchell1988bayesian}, introduces discrete latent variables that directly model the presence or absence of each parameter. This discrete approach is the most direct and accurate representation of a sparsity hypothesis \cite{mohamed2012bayesian}, but the discrete latent space that it imposes is often computationally intractable for models where Bayesian inference is difficult. %that are high-dimensional or dihigh-dimensional models. 

The second approach to Bayesian sparsity uses the scale mixtures of normals \cite{andrews1974scale}, a family of distributions that arise from integrating a zero mean-Gaussian over an unknown variance as
\begin{equation} 
p(\theta) = \int_0^\infty \frac{1}{\sqrt{2\pi}\sigma} \expB{-\frac{\theta^2}{2\sigma^2}} p(\sigma) d\sigma.
\end{equation}
Scale-mixtures of normals can approximate the discrete spike and slab prior by mixing both large and small values of the variance $\sigma^2$. The implicit prior of $L_1$ regularization, the Laplacian, is a member of the scale mixture family that results from an exponentially distributed variance $\sigma^2$. Thus, mixing densities $p(\sigma^2)$ with subexponential tails and more mass near the origin more accurately model sparsity than $L_1$ and are the basis for approaches often referred to as ``Sparse Bayesian Learning'' \cite{tipping2001sparse}. Both the Student-$t$ of Automatic Relevance Determination (ARD) \cite{mackay1994bayesian} and the Horseshoe prior \cite{carvalho2010horseshoe} incorporate these properties.

\begin{figure}[t]
\vskip 0.2in
\begin{center}
\centerline{\includegraphics[width=0.7\columnwidth]{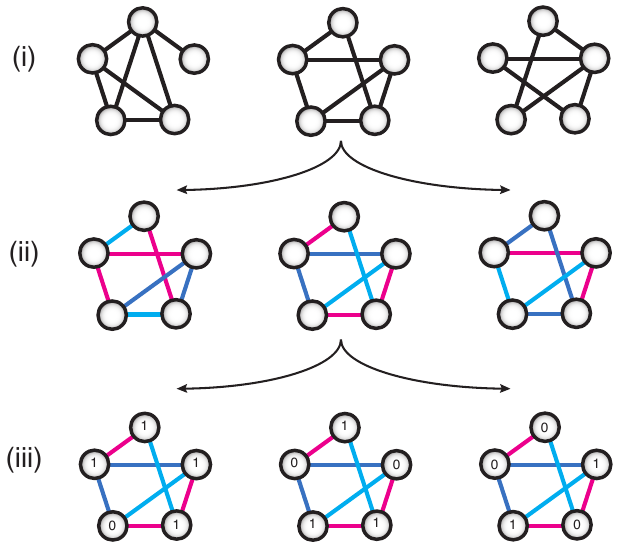}}
\caption{Bayesian inference for discrete undirected graphical models with sparse priors is triply intractable, as the space of possible models spans: (i) all possible sparsity patterns, each of which possesses its own (ii) parameter space, for which every distinct set of parameters has its own (iii) intractable normalizing constant. }
\label{intractable}
\end{center}
\vskip -0.2in
\end{figure}

Applying these favorable, Bayesian approaches to sparsity has been particularly challenging for discrete, undirected models like Boltzmann Machines. Undirected models possess a representational advantage of capturing `collective phenomena' with no directions of causality, but their likelihoods require an intractable normalizing constant \cite{murray2004bayesian}. For a fully observed Boltzmann Machine with $\mathbf x \in \{0,1\}^D$ the distribution\footnote{We exclude biases for simplicity.} is 
\begin{equation} p(\mathbf x | \mathbf J) = \frac{1}{Z(\mathbf J)} \expB{\sum_{i<j} J_{ij} x_i x_j}, \end{equation}
where the partition function $Z(\mathbf J)$ depends on the couplings. Whenever a new set of couplings $\mathbf J$ are considered during inference, the partition function $Z(\mathbf J)$ and corresponding density $p(\mathbf x | \mathbf J)$ must be reevaluated. This requirement for an an intractable calculation embedded within already-intractable nonconjugate inference has led some to term Bayesian learning of undirected graphical models ``doubly intractable'' \cite{murray2006mcmc}. When all $2^{\binom{D}{2}}$ patterns of discrete spike and slab sparsity are added on top of this, we might call this problem ``triply intractable" (Figure \ref{intractable}). Triple-intractability does not mean that this problem is impossible, but it will typically require expensive approaches based on MCMC-within-MCMC \cite{chen2012bayesian}.

Here we present an alternative to MCMC-based approaches for learning undirected models with sparse priors based on stochastic variational inference \cite{hoffman2013stochastic}. We combine three ideas: (i) stochastic gradient variational Bayes \cite{ kingma2013auto, rezende2014stochastic, titsias2014doubly}\footnote{This is also a type of noncentered parameterization, but of the variational distribution rather than the posterior.}, (ii) persistent Markov chains \cite{younes1989parametric}, and (iii) a noncentered parameterization of scale-mixture priors, to inherit the benefits of hierarchical Bayesian sparsity in an efficient variational framework. We make the following contributions:
\begin{itemize}
\item We extend stochastic variational inference to undirected models with intractable normalizing constants by developing a learning algorithm based on \emph{persistent} Markov chains, which we call Persistent Variational Inference (PVI) (Section \ref{PVI_sect}).
\item We introduce a reparameterization approach for variational inference under sparsity-inducing scale-mixture priors (e.g. the Laplacian, ARD, and the Horseshoe) that significantly improves approximation quality by capturing \emph{scale uncertainty} (Section \ref{fadeout_sect}). When combined with Gaussian stochastic variational inference, we call this Fadeout.
\item We demonstrate how a Bayesian approach for learning sparse undirected graphical models with PVI and Fadeout yields significantly improved inferences of both synthetic and real applications in physics and biology (Section \ref{exp_sect}).
\end{itemize}

%%%%%%%%%%%%%%%%%%%%%%%%%%% PVI %%%%%%%%%%%%%%%%%%%%%%%%%%%
\section{Persistent Variational Inference}\label{PVI_sect}

\paragraph{Background: Learning in undirected models} Undirected graphical models, also known as Markov Random Fields, can be written in log-linear form as
\begin{equation}
\label{mrf}
p({\bf x} | \boldsymbol{\theta}) = 
\frac{1}{Z(\boldsymbol{\theta})}\expB{\sum_{i=1}^k \theta_i f_i(\bf x)},
\end{equation}
where $i$ indexes a set of $k$ features $\{f_i({\bf x})\}_{i=1}^k$ and the partition function $Z(\boldsymbol{\theta}) = \sum_{\bf x} \expB{\sum_i \theta_i f_i(\bf x)}$ normalizes the distribution \cite{koller2009probabilistic}. Maximum Likelihood inference selects parameters $\boldsymbol{\theta}$ that maximize the probability of data $\mathcal{D} = \{{\bf x}^{(1)},\ldots, {\bf x}^{(N)}\}$ by ascending the gradient of the (averaged) log likelihood
\begin{equation}
\label{gradmrf}
\frac{\partial}{\partial\theta_i} \frac{1}{N}\log p(\mathcal{D} | \boldsymbol{\theta})  = \mathbb{E}_{\mathcal{D}} \left[f_i({\bf x})\right] - \mathbb{E}_{ p(\mathbf x | \boldsymbol{\theta})} \left[ f_i({\bf x}) \right].
\end{equation}
The first term in the gradient is a data-dependent average of feature $f_i(\bf x)$ over $\mathcal{D}$, while the second term is a data-independent average of feature $f_i(\bf x)$ over the model distribution that often requires sampling \cite{murphy2012machine}\footnote{Depending on the details of the MCMC and the community these approaches are known as Boltzmann Learning, Stochastic Maximum Likelihood, or Persistent Contrastive Divergence \cite{tieleman2008training}.}.

Bayesian learning for undirected models is confounded by the partition function $Z(\boldsymbol{\theta})$. Given the data $\mathcal{D}$, a prior $p(\boldsymbol{\theta})$, and the log  potentials $H[{\bf x}|\boldsymbol{\theta}] = -\sum_i \theta_i f_i(\bf x)$ , the posterior distribution of the parameters is
\begin{equation}
\label{mrf_posterior}
p(\boldsymbol{\theta} | \mathcal{D}) = \frac{p(\boldsymbol{\theta}) \prod_i e^{- H[{\bf x}^{(i)}|\boldsymbol{\theta}]}  / Z(\boldsymbol{\theta}) }{ \int p(\boldsymbol{\theta}')  \prod_i e^{- H[{\bf x}^{(i)}|\boldsymbol{\theta}']}  / Z(\boldsymbol{\theta}') d\boldsymbol{\theta}'},
\end{equation}
which contains an intractable partition function $Z(\boldsymbol{\theta})$ \emph{within} the already-intractable evidence term. As a result, most algorithms for Bayesian learning of undirected models require either doubly-intractable MCMC and/or approximations of the likelihood $p({\bf x} | \boldsymbol{\theta})$.
%Here we will develop an algorithm for directly approximating the posterior $p(\boldsymbol{\theta} | \mathcal{D})$ without approximating the partition function $Z(\boldsymbol{\theta})$ or the likelihood $p({\bf x} | \boldsymbol{\theta})$.
%
\begin{figure*}[t]
\begin{center}
\includegraphics[width=0.6\linewidth]{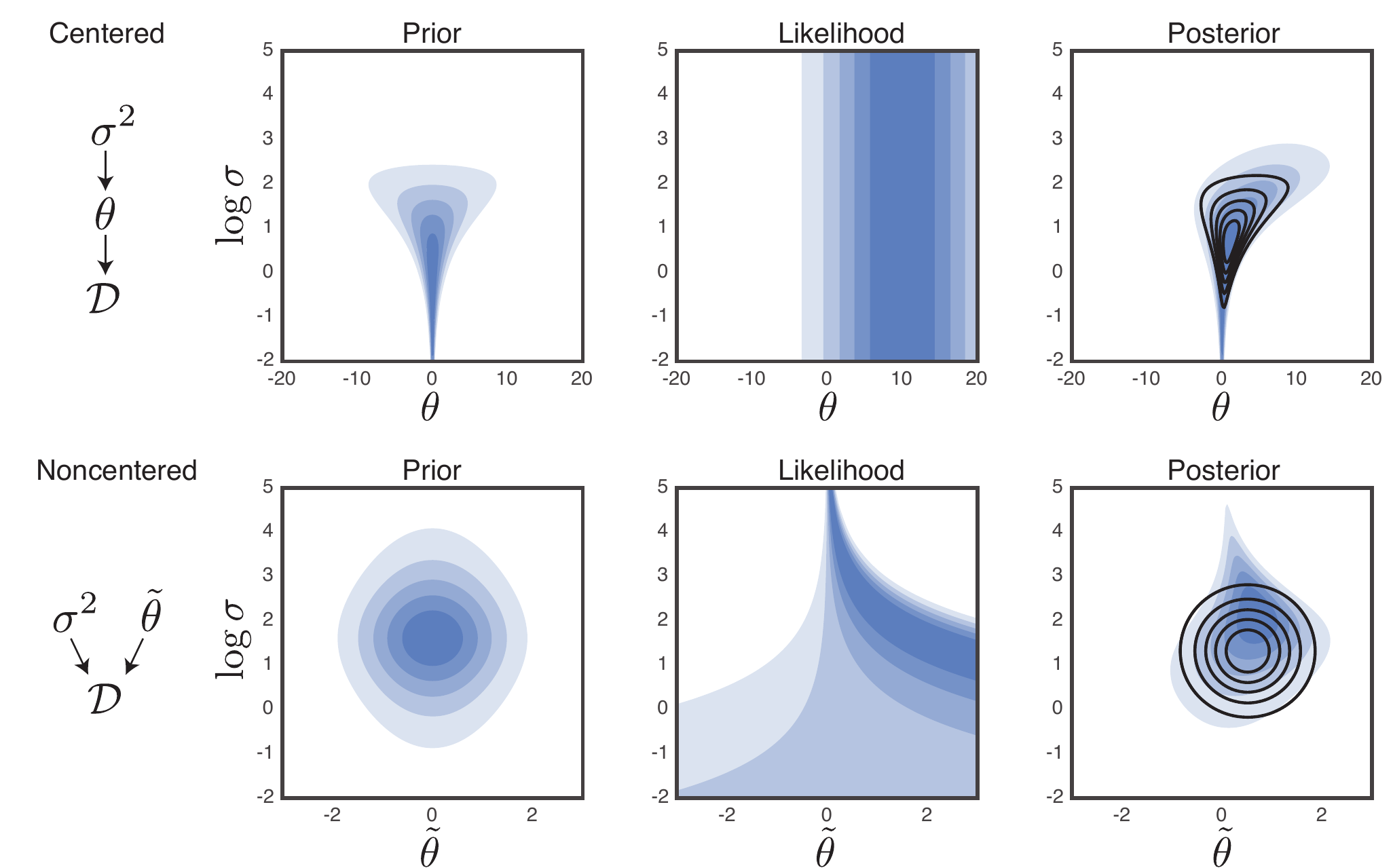}
\caption{Variational inference for sparse priors with noncentered reparameterizations. Several sparsity-inducing priors such as the Laplacian, Student-$t$, and Horseshoe (shown here) can be derived as scale-mixture priors in which each model parameter $\theta$ is drawn from a zero-mean Gaussian with random variance $\sigma^2$ (top row). The dependency of $\theta$ on $\sigma^2$ gives rise to a strongly curved ``funnel" distribution (blue, top left and right) that is poorly modeled by a factorized variational distribution (not shown). A noncentered reparameterization with $\tilde \theta = \theta / \sigma$ trades independence of $\theta$ and $\sigma^2$ in the likelihood (blue, top center) for independence in the prior (blue, bottom left), allowing a factorized variational distribution over noncentered parameters (black contours, bottom right) to implicitly capture the \emph{a priori} correlations between $\theta$ and $\sigma^2$ (black contours, top right). As a result, the variational distribution can more accurately model the bottom of the ``funnel'', which corresponds to sparse estimates.}
\label{fadeout}
\end{center}
\end{figure*}

\paragraph{A tractable estimator for $\nabla$ELBO of undirected models} 
Here we consider how to approximate the intractable posterior in (\ref{mrf_posterior}) without approximating the partition function $Z(\boldsymbol{\theta})$ or the likelihood $p({\bf x} | \boldsymbol{\theta})$ by using variational inference. Variational inference recasts inference with $p(\boldsymbol{\theta} | \mathcal{D})$ as an optimization problem of finding a variational distribution $q(\boldsymbol{\theta} | \boldsymbol{\phi})$ that is closest to $p(\boldsymbol{\theta} | \mathcal{D})$ as measured by KL divergence \cite{jordan1999introduction}. This can be accomplished by maximizing the Evidence Lower BOund
\begin{equation}
\mathcal{L}(\boldsymbol{\phi})  \triangleq \mathbb{E}_{q} \left[ \log p(\mathcal D , \boldsymbol{\theta}) - \log q(\boldsymbol{\theta} | \boldsymbol{\phi}) \right] \leq \log p(\mathcal D).
\end{equation}
For scalability, we would like to optimize the ELBO with methods that can leverage Monte Carlo estimators of the gradient $\nabla_\phi \mathcal{L}(\boldsymbol{\phi})$. One possible strategy for this would be would be to develop an estimator based on the score function \cite{ranganath2014black} with a Monte-Carlo approximation of
\begin{equation}
\label{score}
\nabla_\phi \mathcal{L}  = \mathbb{E}_{q} \left[\nabla_\phi \log q(\boldsymbol{\theta} | \boldsymbol{\phi})\log \frac{p(\mathcal D , \boldsymbol{\theta})}{q(\boldsymbol{\theta} | \boldsymbol{\phi})} \right].
\end{equation}
Naively substituting the likelihood (\ref{mrf}) in the score function estimator (\ref{score}) nests the intractable log partition function $\log Z(\boldsymbol{\theta})$ within the average over $q(\boldsymbol{\theta} | \boldsymbol{\phi})$, making this an untenable (and extremely high variance) approach to inference with undirected models. 

We can avoid the need for a score-function estimator with the `reparameterization trick' \cite{kingma2013auto, rezende2014stochastic, titsias2014doubly} that has been incredibly useful for directed models. Consider a variational approximation $q(\boldsymbol{\theta} | \boldsymbol{\phi}) = \prod_i q(\theta_i | \mu_i , s_i)$ that is a fully factorized (mean field) Gaussian with means $\boldsymbol \mu$ and log standard deviations $\bf s$. The ELBO expectations under $q(\boldsymbol{\theta} | \boldsymbol{\phi})$ can be rewritten as expectations wrt an independent noise source $\boldsymbol\epsilon\sim\mathcal{N}(0, I)$ where\footnote{The $\odot$ operator is an element-wise product.} $\boldsymbol{\theta}(\boldsymbol\epsilon) = \boldsymbol\mu + \expB{\bf s}\odot\boldsymbol\epsilon$.  Then the gradients are
\begin{align}
\label{pathwise1}
\nabla_\mu \mathcal{L}  & = \mathbb{E}_{\boldsymbol\epsilon} \left[ \nabla_\theta \log p(\mathcal D , \boldsymbol{\theta}(\boldsymbol\epsilon))\right], \\
\label{pathwise2}
\nabla_{\bf s} \mathcal{L}  &= \mathbb{E}_{\boldsymbol\epsilon} \left[ \nabla_\theta \log p(\mathcal D , \boldsymbol{\theta}(\boldsymbol\epsilon)) \odot (\boldsymbol{\theta}(\boldsymbol\epsilon) - \boldsymbol\mu)\right] + {\bf 1}.
\end{align}
Because these expectations require only the gradient of the likelihood $\nabla_\theta\log p(\mathcal D |\boldsymbol{\theta})$, the gradient for the undirected model (\ref{gradmrf}) can be substituted to form a nested expectation for $\nabla_\phi \mathcal{L}(\boldsymbol{\phi})$. This can then be used as a Monte Carlo gradient estimator by sampling $\boldsymbol\epsilon\sim\mathcal{N}(0, I), {\bf x} \sim p({\bf x} | \boldsymbol{\theta}(\boldsymbol\epsilon))$.
%\begin{align}
%\nabla_\mu \mathcal{L}  & = \mathbb{E}_{\boldsymbol\epsilon} \left[ \mathbb{E}_{\mathcal{D}} \left[f_i({\bf x})\right] - \mathbb{E}_{ p(\mathbf x | \boldsymbol{\theta})} \left[ f_i({\bf x}) \right] + \log p(\boldsymbol{\theta}(\boldsymbol{\epsilon}))\right], \\
%\nabla_{\bf s} \mathcal{L}  &= \mathbb{E}_{\boldsymbol\epsilon} \left[ \nabla_\theta \log p(\mathcal D , \theta(\boldsymbol\epsilon)) \odot (\theta(\boldsymbol\epsilon) - \boldsymbol\mu)\right] + {\bf 1}.
%\end{align}

\paragraph{Persistent gradient estimation} In Stochastic Maximum Likelihood estimation for undirected models, the intractable gradients of (\ref{gradmrf}) are estimated by sampling $p({\bf x} | \boldsymbol{\theta})$. Although sampling-based approaches are slow, they can be made considerably more efficient by running a set of Markov chains in parallel with state that \emph{persists} between iterations \cite{younes1989parametric}. Persistent state maintains the Markov chains near their equilibrium distributions, which means that they can quickly re-equilibrate after perturbations to the parameters $\boldsymbol{\theta}$ during learning.

We propose variational inference in undirected models based on persistent gradient estimation of $\nabla_\theta \log p(\mathcal{D} | \boldsymbol{\theta})$ and refer to this as Persistent Variational Inference (PVI) (Algorithm in Appendix). Following the notation of PCD-$n$ \cite{tieleman2008training}, PVI-$n$ refers to using $n$ sweeps of Gibbs sampling with persistent Markov chains between iterations. This approach is generally compatible with any estimators of $\nabla$\textsc{ELBO} that are based on the gradient of the log likelihood, several examples of which are explained in \cite{kingma2013auto, rezende2014stochastic, titsias2014doubly}.

\paragraph{Behavior of the solution for Gaussian $q$} When the variational approximation is a fully factorized Gaussian $q(\boldsymbol{\theta} | \boldsymbol{\mu}, \boldsymbol{\sigma})$ and the prior is flat $p(\boldsymbol{\theta}) \propto 1$, the solution to $\boldsymbol{\mu}^\star, \boldsymbol{\sigma}^\star = \arg\max_{\boldsymbol{\mu}, \boldsymbol{\sigma}} \mathcal{L}(\boldsymbol{\mu}, \boldsymbol{\sigma})$ will satisfy
\begin{equation}
\mathbb{E}_{\mathcal{D}} \left[f_i({\bf x})\right] = \mathbb{E}_{\tilde p} \left[f_i({\bf x})\right], ~~~ \sigma_i^\star = \frac{1}{N ~ \mathbb{E}_{\tilde p} \left[\epsilon_i f_i({\bf x})\right]}
\end{equation}
where $\tilde p = p({\bf x} | \theta(\epsilon))p(\epsilon)$ is an extended system of the original undirected model in which the parameters $\theta_i = \mu_i + \epsilon_i \sigma_i$ fluctuate according to the variational distribution. This bridges to the Maximum Likelihood solution as $N \to \infty$ and $\sigma_i^\star \to 0$, while accounting for uncertainty in the parameters at finite sample sizes with the inverse of `sensitivity' $\mathbb{E}_{\tilde p} \left[ \epsilon_i f_i({\bf x})\right]$.

%%%%%%%%%%%%%%%%%%%%%%%%%%% FADEOUT %%%%%%%%%%%%%%%%%%%%%%%%%%%
\section{Fadeout}\label{fadeout_sect}

\begin{table}[b]
\caption{Common priors as scale-mixtures of normal distributions}
\begin{center}
\begin{tabular}{ccc}
\hline
Prior & Hyperprior & $p(\log \sigma)$  \vspace*{1mm}\\
\hline \vspace*{-3mm}\\
Gaussian ($L_2$) & $\sigma^2 = \frac{1}{2\lambda}$ \vspace*{1mm} & constant \\
Laplacian ($L_1$) &  $\sigma^2 \sim \text{Exponential}$ & $2 \lambda  e^{-\lambda  \sigma ^2} \sigma ^2$ \vspace*{1mm}\\
Student-$t$ (ARD) & $\sigma^2 \sim \text{Inv. Gamma}$ & \footnotesize$\frac{2 \beta ^{\alpha }}{\Gamma (\alpha )} e^{-\frac{\beta }{\sigma ^2}} \sigma ^{-2 \alpha }$ \vspace*{1mm}\\
Horseshoe & $\sigma \sim \text{Half-Cauchy}$ & $\frac{2s}{\pi} \frac{\sigma}{s^2 + \sigma^2} $ \vspace*{1mm}\\
\hline
\end{tabular}
\end{center}
\label{mixtures}
\end{table}%

\subsection{Noncentered Parameterizations of Hierarchical Priors}
Hierarchical models are powerful because they impose \emph{a priori} correlations between latent variables that reflect problem-specific knowledge. For scale-mixture priors that promote sparsity, these correlations come in the form of \emph{scale uncertainty}. Instead of assuming that the scale of a parameter in a model is known \emph{a priori}, we posit that it is normally distributed with a randomly distributed variance $p(\sigma^2)$. The joint prior $p(\theta|\sigma^2)p(\sigma^2)$ gives rise to a strongly curved `funnel' shape (Figure \ref{fadeout}) that illustrates a simple but profound principle about hierarchical models: as the hyperparameter $\log \sigma$ decreases and the prior accepts a smaller range of values for $\theta$, normalization increases the probability density at the origin, favoring sparsity. This normalization-induced sharpening has been called called a Bayesian Occam's Razor \cite{mackay2003information}.

While normalization-induced sharpening gives rise to sparsity, these extreme correlations are a disaster for mean-field variational inference. Even if a tremendous amount of probability mass is concentrated at the base of the funnel, an uncorrelated mean-field approximation will yield estimates near the top. The result is a potentially non-sparse estimate from a very-sparse prior.

The strong coupling of hierarchical funnels also plagues exact methods based on MCMC with slow mixing, but the statistics community has found that these geometry pathologies can be effectively managed by transformations. Many models can be rewritten in a noncentered form where the parameters and hyperparmeters are \emph{a priori} independen \cite{papaspiliopoulos2007general, betancourt2013hamiltonian}. For the scale-mixtures of normals, this change of variables is
\begin{equation} \left\{ \theta, \log \sigma \right\} \to \left\{\frac{\theta}{\sigma}, \log\sigma \right\} \end{equation}

\begin{algorithm}[t]
   \caption{Computing $\nabla \textsc{\footnotesize ELBO}$ for Fadeout}
   \label{Fadeout}
   \begin{algorithmic}
   \STATE {\bfseries Require:} Global parameters $\{\mu_{\tau}, s_{\tau}\}$
   \STATE {\bfseries Require:} Local parameters $\{\mu_{\tilde \theta}, \mu_{\log\boldsymbol\sigma}, s_{\tilde \theta}, s_{\log\boldsymbol\sigma}\}$
   \STATE {\bfseries Require:} Hyperprior gradient $\nabla_{\log\boldsymbol\sigma, \boldsymbol \tau} \log p(\log\boldsymbol\sigma , \boldsymbol \tau)$
   \STATE {\bfseries Require:} Likelihood gradient $\nabla_\theta p(\mathbf x| \boldsymbol \theta)$
   \STATE // \emph{Sample from variational distribution}
   \STATE $\mathbf z_1 \sim \mathcal{N}(0, I_{|\boldsymbol\tau|})$, $\mathbf z_2 \sim \mathcal{N}(0, I_{|\boldsymbol{\tilde\theta}|})$, $\mathbf z_3 \sim \mathcal{N}(0, I_{|\boldsymbol\sigma|})$
   \STATE $ \boldsymbol\tau \gets \mu_{\tau} + \exp\{s_{\tau}\} \odot \mathbf z_1$
   \STATE $ \boldsymbol{\tilde\theta} \gets \mu_{\tilde \theta} + \exp\{s_{\tilde \theta}\} \odot \mathbf z_2$
   \STATE $ \boldsymbol\sigma \gets \expB{\mu_{\log\sigma} + \expB{s_{\log\sigma}} \odot \mathbf z_3}$
    \STATE $ \boldsymbol{\theta} \gets \boldsymbol{\tilde\theta} \odot \boldsymbol\sigma$
    \STATE // \emph{Centered global parameters}
   \STATE $\nabla_{\mu_{\boldsymbol \tau}} \mathcal{L} \gets \nabla_{\boldsymbol \tau} \log p (\log \sigma , \boldsymbol \tau)$
    \STATE $\nabla_{s_{\boldsymbol \tau}} \mathcal{L} \gets \expB{s_{\boldsymbol \tau}} \odot \mathbf z_1 \odot\nabla_{\mu_{\boldsymbol \tau}} \mathcal{L}  + 1$
   \STATE // \emph{Noncentered local parameters}
   \STATE $\nabla_{\mu_{\tilde \theta}} \mathcal{L} \gets \boldsymbol\sigma \odot \nabla_\theta \log p(\boldsymbol x |\boldsymbol \theta) - \boldsymbol{\tilde\theta}$
   \STATE $\nabla_{\mu_{\log\sigma}} \mathcal{L} \gets \boldsymbol\theta \odot \nabla_\theta \log p(\boldsymbol x |\boldsymbol \theta)  + \nabla_{\log\sigma} \log p (\log \sigma , \boldsymbol \tau)$
   \STATE $\nabla_{s_{\tilde \theta}} \mathcal{L} \gets \expB{s_{\tilde \theta}} \odot \mathbf z_2 \odot\nabla_{\mu_{\tilde \theta}} \mathcal{L}  + 1$
   \STATE $\nabla_{s_{\log\sigma}}  \mathcal{L} \gets  \expB{s_{\log\sigma}} \odot  \mathbf z_3 \odot\nabla_{\mu_{\log\sigma}} \mathcal{L} + 1$
\end{algorithmic}
\end{algorithm}

Then $\tilde\theta \triangleq \frac{\theta}{\sigma} \sim N(0,1)$ while preserving $\tilde\theta \sigma \sim N(0,\sigma^2)$. In noncentered form, the joint prior is independent and well approximated by a mean-field Gaussian, while the likelihood will be variably correlated depending on the strength of the data (Figure \ref{fadeout}). In this sense, centered parameterizations (CP) and noncentered parameterizations (NCP) are usually framed as favorable in strong and weak data regimes, respectively.\footnote{Although ``weak data" may seem unrepresentative of typical problems in machine learning, it is important to remember that a sufficiently large and expressive model can make most data weak.} 

We propose the use of non-centered parameterizations of scale-mixture priors for mean-field Gaussian variational inference. For convenience, we like to call this Fadeout (see next section). Fadeout can be easily implemented by either (i) using the chain rule to derive the gradient of the Evidence Lower BOund (\textsc{\footnotesize ELBO}) (Algorithm \ref{Fadeout}) or, for differentiable models, (ii) rewriting models in noncentered form and using automatic differentiation tools such as Stan \cite{kucukelbir2017automatic} or \texttt{autograd}\footnote{github.com/HIPS/autograd} for ADVI. The only two requirements of the user are the gradient of the likelihood function and a choice of a global hyperprior, several options for which are presented in Table \ref{mixtures}.

\paragraph{Estimators for the centered posterior.} Fadeout optimizes a mean-field Gaussian variational distribution over the noncentered parameters $q(\boldsymbol {\tilde \theta}, 
\log \boldsymbol \sigma)$. As an estimator for the centered parameters, we use the mean-field property to compute the centered posterior mean as $\mathbb{E}_q [\theta] = \mathbb{E}_q [\boldsymbol {\tilde \theta}] \odot \mathbb{E}_q [\boldsymbol \sigma]$, giving \footnote{The term $\frac{ 1}{2}e^{2 s_{\log \boldsymbol \sigma}}$ is optional in the sense that including it corresponds to averaging over the hyperparameters, whereas discarding it corresponds to optimizing the hyperparameters (Empirical Bayes). We included it for all experiments.}
\begin{equation}
\hat {\theta} = \boldsymbol \mu_{\tilde \theta} \odot \expB{\boldsymbol\mu_{\log \boldsymbol \sigma} + \frac{ 1}{2}e^{2 s_{\log \boldsymbol \sigma}} }
\end{equation}

\begin{figure}[t]
\begin{center}
\includegraphics[width=0.9\linewidth]{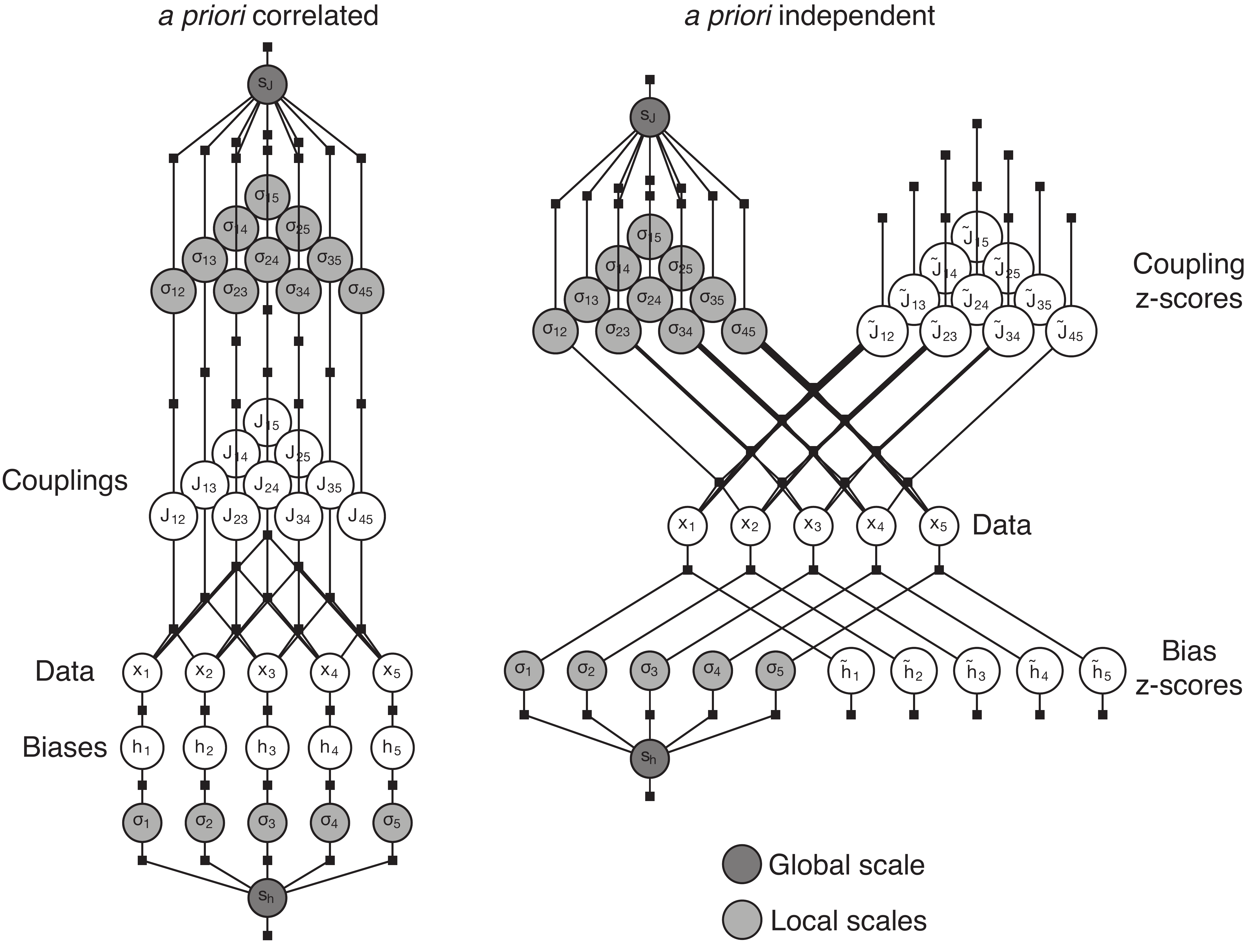}
\caption{An undirected model with a scale mixture prior (factor graph on left) can be given \emph{a priori} independence of the latent variables by a noncentered parameterization (factor graph on right). This is advantageous for mean-field variational inference that imposes \emph{a posteriori} independence.
}
\label{model}
\end{center}
\end{figure}
\begin{figure}[t]
\centering
\includegraphics[width=0.9\linewidth]{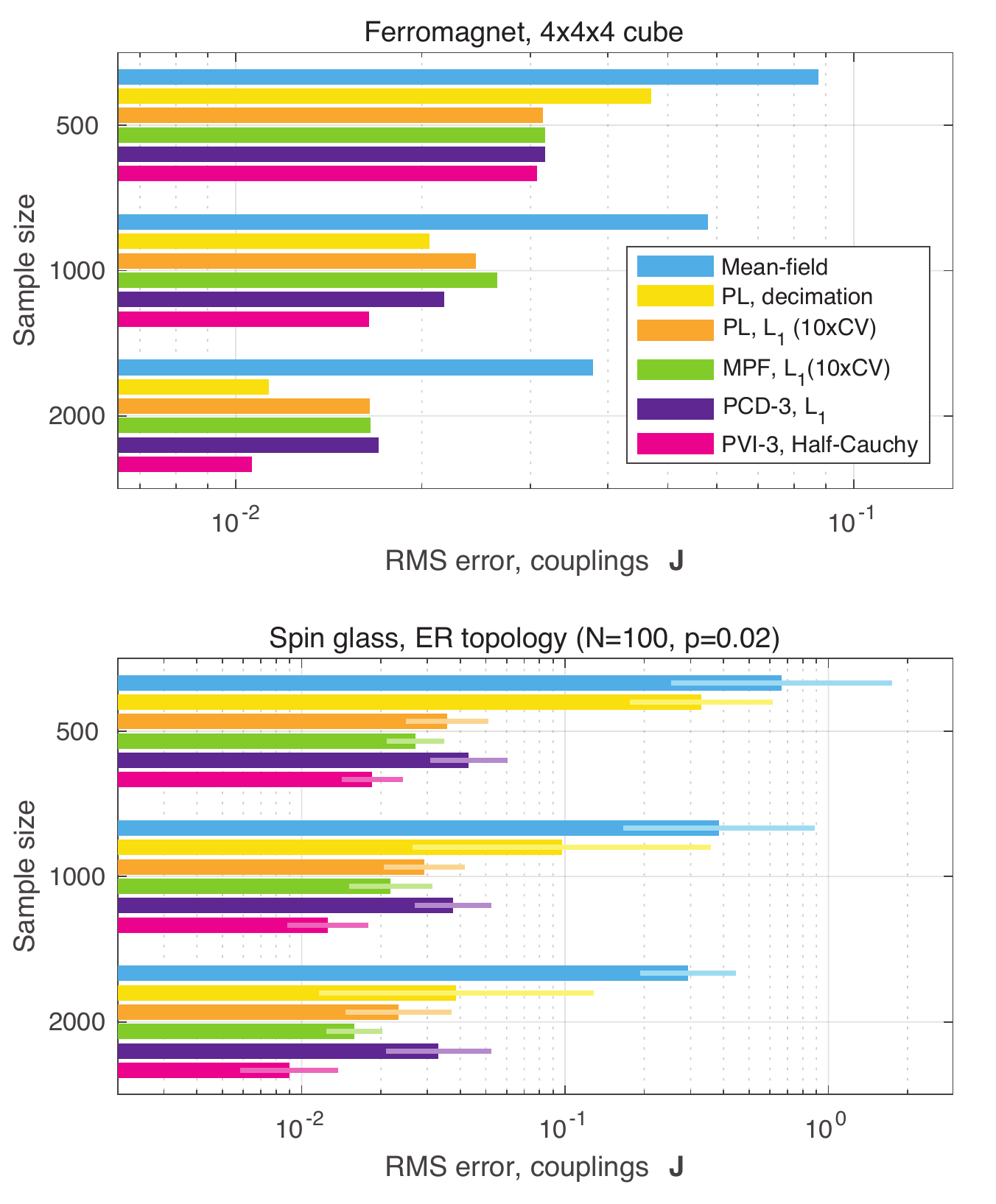}%
\caption{Inverse Ising. Combining Persistent VI with a noncentered Horseshoe prior (Half-Cauchy hyperprior) attains lower error on simulated Ising systems than standard methods for point estimation including: Pseudolikelihood (PL) with $L_1$ or decimation regularization \cite{schmidt2010graphical, aurell2012inverse, decelle2014pseudolikelihood}, Minimum Probability Flow (MPF) \cite{sohl2011new}, and Persistent Contrastive Divergence (PCD) \cite{tieleman2008training}. For the spin glass, error bars are two logarithmic standard deviations across 5 simulated systems.
\label{spinglass}} 
\end{figure}
%

% SyntheticProtein
\begin{figure*}[t]
\begin{center}
\includegraphics[width=\linewidth]{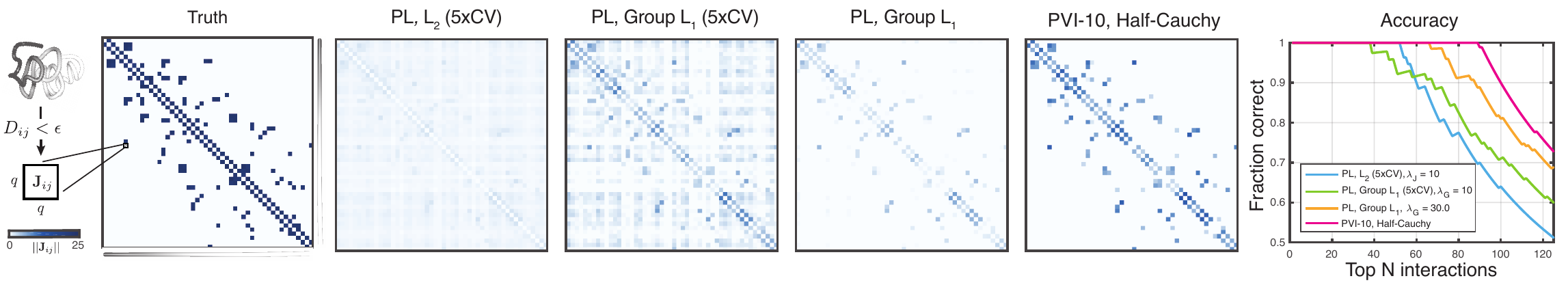}
\vspace*{-5mm}
\caption{Synthetic protein. For reconstructing interactions in a synthetic 20-letter spin-glass, a hierarchical Bayesian approach based on Persistent VI and a noncentered group Horseshoe prior (Half-Cauchy hyperprior) identifies true interactions with more accuracy and less shrinkage than Group $L_1$. Each $i,j$ pair is the norm of a $20\times20$ factor coupling the amino acid at position $i$ to the amino acid at position $j$.}
\label{potts}
\end{center}
\end{figure*}

\subsection{Connection to Dropout}
Dropout regularizes neural networks by perturbing hidden units in a directed network with multiplicative Bernoulli or Gaussian noise \cite{srivastava2014dropout}. Although it was originally framed as a heuristic, Dropout has been subsequently interpreted as variational inference under at least two different schemes \cite{gal2016dropout, kingma2015variational}. Here, we interpret Fadeout the reverse way, where we introduced it as variational inference and now notice that it looks similar to lognormal Dropout.\footnote{Rather than attempting to explain Dropout, the intent is to lend intuition about noncentered scale-mixture VI.} If we take the uncertainty in $\boldsymbol{\tilde\theta}$ as low and clamp the other variational parameters, the gradient estimator for Fadeout is:
\begin{align*}
\mathbf z &\sim \mathcal{N}(0, I_{|\theta|}) \\
\boldsymbol\sigma &\gets \expB{\mu_{\log\sigma} + \expB{s_{\log\sigma}} \odot \mathbf z} \\
\boldsymbol{\theta} &\gets \boldsymbol{\tilde\theta}  \expB{\mu_{\log\sigma} + \expB{s_{\log\sigma}} \odot \mathbf z} \\
\nabla_{\mu_{\tilde \theta}} \mathcal{L} &\gets \boldsymbol\sigma \odot \nabla_\theta \log p(\boldsymbol x |\theta) - \boldsymbol{\tilde\theta}
\end{align*}
This is the gradient estimator for a lognormal version of Dropout with an $L_2$ weight penalty of $\frac{1}{2}$. At each sample from the variational distribution, Fadeout introduces \emph{scale noise} rather than the Bernoulli noise of Dropout. The connection to Dropout would seem to follow naturally from the common interpretation of scale mixtures as continuous relaxations of spike and slab priors \cite{engelhardt2014bayesian} and the idea that Dropout can be related to variational spike and slab inference \cite{louizos2015smart}.

%%%%%%%%%%%%%%%%%%%%%%%%%%% Experiments %%%%%%%%%%%%%%%%%%%%%%%%%%%
\section{Experiments}\label{exp_sect}

\subsection{Physics: Inferring Spin Models}

\paragraph{Ising model} The Ising model is a prototypical undirected model for binary systems that includes both pairwise interactions and (potentially) sitewise biases. It can be seen as the fully observed case of the Boltzmann machine, and is typically parameterized with signed spins ${\bf x} \in \{-1,1\}^D$ and a likelihood given by
\begin{equation} \small p(\mathbf x | \mathbf h, \mathbf J) = \frac{1}{Z(\mathbf h,\mathbf J)} \exp \left\{\sum_i h_i x_i + \sum_{i<j} J_{ij} x_i x_j \right\}. \end{equation}

Originally proposed as a minimal model of how long range order arises in magnets, it continues to find application in physics and biology as a model for phase transitions and quenched disorder in spin glasses \cite{nishimori2001statistical} and collective firing patterns in neural spike trains \cite{schneidman2006weak, shlens2006structure}.

\paragraph{Hierarchical sparsity prior} One appealing feature of the Ising model is that it allows a sparse set of underlying couplings $\bf J$ to give rise to long-range, distributed correlations across a system. Since many physical systems are thought to be dominated by a small number of relevant interactions, $L_1$ regularization has been a favored approach for inferring Ising models. Here, we examine how a more accurate model of sparsity based on the Horseshoe prior (Figure \ref{model}) can improve inferences in these systems. Each coupling $J_{ij}$ and bias parameter $h_i$ is given its own scale parameter which are in turn tied under a global Half-Cauchy prior for the scales (Figure \ref{model}, Appendix).

\paragraph{Simulated datasets} We generated synthetic couplings for two kinds of Ising systems: (i) a slightly sub-critical cubic ferromagnet ($J_{ij} > 0$ for neighboring spins) and (ii) a Sherrington-Kirkpatrick spin glass diluted on an Erd\"{o}s-Renyi random graph with average degree 2. We sampled synthetic data for each system with the Swendsen-Wang algorithm (Appendix) \cite{swendsen1987nonuniversal}.

\paragraph{Results} On both the ferromagnet and the spin glass, we found that Persistent VI with a noncentered Horseshoe prior (Fadeout) gave estimates with systematically lower reconstruction error of the couplings $\bf J$ (Figure \ref{spinglass}) versus a variety of standard methods in the field (Appendix).
\subsection{Biology: Reconstructing 3D Contacts in Proteins from Sequence Variation}\label{potts_sect}

\begin{figure*}[t]
\begin{center}
\includegraphics[width=\textwidth]{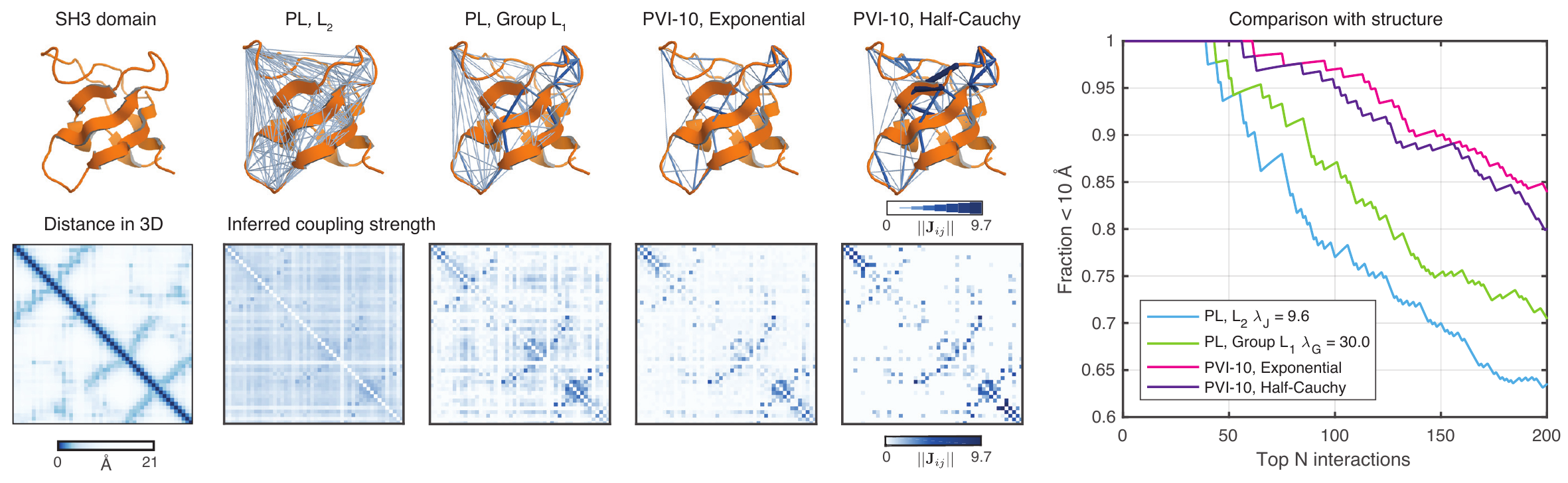}
\caption{Unsupervised protein contact prediction. When inferring a pairwise undirected model for protein sequences in the SH3 domain family, hierarchical Bayesian approaches based on Persistent VI and noncentered scale mixture priors (Half-Cauchy for Group Horseshoe and Exponential for a Multivariate Laplace) identify local interactions that are close in 3D structure without tuning parameters. When group $L_1$-regularized maximum Pseudolikelihood estimation is tuned to give the same largest effect size as the Multivariate Laplace, the hierarchical approaches based on Persistent VI are more predictive of 3D proximity (right).}
\label{realprotein}
\end{center}
\end{figure*}

\paragraph{Potts model} The Potts model generalizes the Ising model to non-binary categorical data. The factor graph is the same (Figure \ref{model}), except each spin $x_i$ can adopt $q$ different categories with ${\bf x} \in \{1,\ldots,q\}^D$ and each $\bf J_{ij}$ is a $q\times q$ matrix as
\begin{equation}
\footnotesize
p({\bf x} | {\mathbf h}, {\mathbf J} ) = \frac{1}{Z(\mathbf h, \mathbf J)} \exp\left\{\sum_{i} h_i(x_i) + \sum_{i < j} J_{ij}(x_i, x_j) \right\}.
\end{equation}
The Potts model has recently generated considerable excitement in biology, where it has been used to infer 3D contacts in biological molecules solely from patterns of correlated mutations in the sequences that encode them \cite{marks2011protein, morcos2011direct}. These contacts are have been sufficient to predict the 3D structures of proteins, protein complexes, and RNAs \cite{marks2012protein}.

\paragraph{Group sparsity} Each pairwise factor ${\bf J}_{ij}$ in a Potts model contains $q\times q$ parameters capturing all possible joint configurations of $x_i$ and $x_j$. One natural way to enforce sparsity in a Potts model is at the level of each $q\times q$ group. This can be accomplished by introducing a single scale parameter $\sigma_{ij}$ for all $q\times q$ z-scores ${\bf \tilde J}_{ij}$. We adopt this with the same Half-Cauchy hyperprior as the Ising problem, giving the same factor graph (Figure \ref{model}) now corresponding to a Group Horseshoe prior \cite{hernandez2013generalized}. In the real protein experiment, we also consider an exponential hyperprior, which corresponds to a Multivariate Laplace distribution \cite{eltoft2006multivariate} over the groups.

\paragraph{Synthetic protein data} We first investigated the performance of Persistent VI with group sparsity on a synthetic protein experiment. We constructed a synthetic Potts spin glass with a topology inspired by biological macromolecules. We generated synthetic parameters based on contacts in a simulated polymer and sampled 2000 sequences with  $2\times 10^6$ steps of Gibbs sampling (Appendix).

\paragraph{Results for a synthetic protein} We inferred couplings with 400 of the sampled sequences using PVI with group sparsity and two standard methods of the field: $L_2$ and Group $L_1$ regularized maximum pseudolikelihood (Appendix). PVI with a noncentered Horseshoe yielded more accurate (Figure \ref{potts}, right), less shrunk (Figure \ref{potts}, left) estimates of interactions that were more predictive of the 1600 remaining test sequences (Table \ref{testpseudo}). The ability to generalize well to new sequences will likely be important to the related problem of predicting mutation effects with unsupervised models of sequence variation \cite{hopf2017mutation,figliuzzi2015coevolutionary}.

\begin{table}[b]
\vskip -0.3in
\caption{Average log-pseudolikelihood for test sequences.}
\begin{center}
\begin{small}
%\begin{sc}
\begin{tabular}{lccr}
\hline
Method & $ -\log \text{PL}({\bf x}|{\bf h},{\bf J})$ & Runtime (s) \\
\hline
PL, $L_2$ (5xCV)    & 67.3 & 375 \\
PL, Group $L_1$ (5xCV) & 59.6 & 303\\
PVI-3, Half-Cauchy & 54.2 & 585\\
\hline
\end{tabular}
\label{testpseudo}
%\end{sc}
\end{small}
\end{center}
\vskip -0.1in
\end{table}

\paragraph{Results for natural sequence variation} We applied the hierarchical Bayesian model from the protein simulation to model across-species amino acid covariation in the SH3 domain family (Figure \ref{realprotein}). Transitioning from simulated to real protein data is particularly challenging for Bayesian methods because available sequence data are highly non-independent due to a shared evolutionary history. We developed a new method for estimating the effective sample size (Appendix) which, when combined standard sequence reweighting techniques, yielded a reweighted effective sample size of 1,012 from 10,209 sequences.

The hierarchical Bayesian approach gave highly localized, sparse estimates of interactions compared to the two predominant methods in the field, $L_2$ and group $L_1$ regularized pseudolikelihood (Figure \ref{realprotein}). When compared to solved 3D structures for SH3 (Appendix), we found that the inferred interactions were considerably more accurate at predicting amino acids close in structure. Importantly, the hierarchical Bayesian approach accomplished this inference of strong, accurate interactions without a need to prespecify hyperparameters such as $\lambda$ for $L_2$ or $L_1$ regularization. This is particularly important for natural biological sequences because the non-independence of samples limits the utility of cross validation for setting hyperparameters.

%%%%%%%%%%%%%%%%%%%%%%%%%%% Related Work %%%%%%%%%%%%%%%%%%%%%%%%%%%
\section{Related work}

\subsection{Variational Inference}
One strategy for improving variational inference is to introduce correlations in variational distribution by geometric transformations. This can be made particularly powerful by using backpropagation to learn compositions of transformations that capture the geometry of complex posteriors \cite{rezende2015variational,tran2015variational}. Noncentered parameterizations of models may be complementary to these approaches by enabling more efficient representations of correlations between parameters and hyperparameters.

Most related to this work, \cite{louizos2017bayesian,ghosh2017model} show how variational inference with noncentered scale-mixture priors can be useful for Bayesian learning of neural networks, and how group sparsity can act as a form of automatic compression and model selection.

\subsection{Maximum Entropy}
Much of the work on inference of undirected graphical models has gone under the name of the Maximum Entropy method in physics and neuroscience, which can be equivalently formulated as maximum likelihood in an exponential family \cite{mackay2003information}. From this maximum likelihood interpretation, $L_1$ regularized-maximum entropy modeling (MaxEnt) corresponds to the disfavored ``integrate-out" approach to inference in hierarchical models\footnote{To see this, note that $L_1$-regularized MAP estimation is equivalent to integrating out a zero-mean Gaussian prior with unknown, exponentially-distributed variance} \cite{mackay1996hyperparameters} that will introduce significant biases to inferred parameters \cite{macke2011biased}. One solution to this bias was foreshadowed by methods for estimating entropy and Mutual Information, which used hierarchical priors to integrate over a large range of possible model complexities \cite{nemenman2002entropy, archer2013bayesian}. These hierarchical approaches are favorable because in traditional MAP estimation any top level parameters that are fixed before inference (e.g. a global pseudocount $\alpha$) introduce strong constraints on allowed model complexity. The improvements from PVI and Fadeout may be seen as extending this hierarchical approach to full systems of discrete variables.

%%%%%%%%%%%%%%%%%%%%%%%%%%% Conclusion %%%%%%%%%%%%%%%%%%%%%%%%%%%
\section{Conclusion}
We introduced a framework for scalable Bayesian sparsity for undirected graphical models composed of two methods. The first is an extension of stochastic variational inference to work with undirected graphical models that uses persistent gradient estimation to bypass estimating partition functions. The second is a variational approach designed to match the geometry of hierarchical, sparsity-promoting priors. We found that, when combined, these two methods give substantially improved inferences of undirected graphical models on both simulated and real systems from physics and computational biology.

\section*{Acknowledgements} 
We thank David Duvenaud, Finale Doshi-Velez, Miriam Huntley, Chris Sander, and members of the Marks lab for helpful comments and discussions. JBI was supported by a NSF Graduate Research Fellowship DGE1144152 and DSM by NIH grant 1R01-GM106303. Portions of this work were conducted on the Orchestra HPC Cluster at Harvard Medical School.

\bibliography{sparsity} 
\bibliographystyle{icml2017}

\section{Appendix I: PVI algorithm}

\begin{algorithm*}[htbp]
   \label{PVI}
   \caption{Persistent Variational Inference (PVI-$n$) with Gaussian $q(\theta|\phi)$}
   \begin{algorithmic}
       \STATE {\bfseries Require:} Model. Undirected $p({\bf x} | \boldsymbol \theta)$ defined by $k$ features $\{f_i(\bf x)\}_{i=1}^k$ on ${\bf x} \in \{1,\ldots,q \}^D$
       \STATE {\bfseries Require:} Data. Expectations of the features $\{\mathbb{E}_{\mathcal{D}} \left[f_i({\bf x})\right]\}_{i=1}^k$ and sample size $N$
       \STATE {\bfseries Require:} Prior. Prior gradient $\nabla \log P({\boldsymbol \theta})$
       \STATE {\bfseries Require:} Number of Gibbs sweeps $n$, Markov Chains $M$, variational samples $Q$
       \STATE {\bfseries Require:} Initial variational parameters ${\boldsymbol \mu}_0, \log{\boldsymbol \sigma}_0$ (e.g. $\{0, -3\}$)
       \STATE // \emph{Initialize parameters and Markov chains $\tilde x$}
       \STATE ${\boldsymbol \mu} \gets {\boldsymbol \mu}_0$, ${\log\boldsymbol \sigma} \gets \log{\boldsymbol \sigma}_0$
       \STATE ${\bf \tilde x}^{(1:M)} \gets \textrm{RandInt}(1,q)$
       \STATE $t \gets 0$
       \WHILE{not converged}
       		\STATE // \emph{Estimate $\nabla \textsc{\footnotesize ELBO}$ with $Q$ samples from the variational distribution}
			\STATE $\nabla_{\boldsymbol \mu} \mathcal{L} \gets 0$, $\nabla_{\log\boldsymbol \sigma} \mathcal{L} \gets 0$
			\FOR{$s = 1 \ldots Q$} 
				%\STATE // \emph{Sample from the variational distribution (Gaussian)}
				\STATE ${\mathbf {\boldsymbol \epsilon}} \sim \mathcal{N}(0, I_{|\boldsymbol\mu|})$
				\STATE ${\boldsymbol \theta} \gets {\boldsymbol \mu} + {\boldsymbol \sigma} \odot {\boldsymbol \epsilon}$
				\STATE // \emph{Estimate model-dependent expectations ${\bf E}$, where $E_i = \mathbb{E}_{ p(\mathbf x | \boldsymbol \theta)} \left[ f_i({\bf x}) \right]$}
				\STATE $ {\bf E} \gets {\boldsymbol 0}$
					\FOR{$m = 1 \ldots M$}
						\FOR{$j = 1 \ldots n$}
							\STATE ${\bf \tilde x}^{(m)} \gets \operatorname{GibbsSweep}(p({\bf x} | \boldsymbol \theta), {\bf \tilde x}^{(m)})$
							 
							\STATE ${\bf E} \gets {\bf E} + \frac{1}{M n}\{ f_i({\bf \tilde x}^{(m)}) \}_{i=1}^k$
						\ENDFOR
					\ENDFOR
				\STATE // \emph{Compute stochastic gradient}
				\STATE ${\bf G} \gets N (\mathbb{E}_{\mathcal{D}} \left[f_i({\bf x})\right] - {\bf E}) + \nabla \log P({\boldsymbol \theta})$
				\STATE $\nabla_{\boldsymbol \mu} \mathcal{L} \gets \nabla_{\boldsymbol \mu} \mathcal{L} + \frac{1}{Q} {\bf G}$
				\STATE $\nabla_{\log{\boldsymbol \sigma}} \mathcal{L} \gets \nabla_{\log{\boldsymbol \sigma}} \mathcal{L} + \frac{1}{Q} \left({\bf G} \odot \left({\boldsymbol \theta - \boldsymbol \mu} \right) + 1\right)$
			\ENDFOR
			\STATE // \emph{Update parameters with Robbins-Monro sequence $\{\rho_t\}$}
			\STATE ${\boldsymbol \mu} \gets {\boldsymbol \mu} + \rho_t \nabla_{\boldsymbol \mu} \mathcal{L}$
			\STATE ${\log\boldsymbol \sigma} \gets {\log\boldsymbol \sigma} + \rho_t \nabla_{\log{\boldsymbol \sigma}}$
			\STATE $t \gets t + 1$
	   \ENDWHILE
	\end{algorithmic}
\end{algorithm*}

See Algorithm \ref{PVI}.

\section{Appendix II: Experiments}
\subsection{Spin Models}
We generated two synthetic systems. The first system was ferromagnetic (all $J\geq0$) with 64 spins, where neighboring spins $x_i$, $x_j$ have a nonzero interaction of $J_{ij} = 0.2$ if adjacent on a $4\times4\times4$ periodic lattice. This coupling strength equates to being slightly above the critical temperature, meaning the system will be highly correlated despite the underlying interactions being only nearest-neighbor.

The second system was a diluted Sherrington-Kirkpatrick spin glass \cite{sherrington1975solvable, aurell2012inverse} with 100 spins. The couplings in this model were defined by Erd\H{o}s-Renyi random graphs \cite{erdos1960evolution} with non-zero edge weights distributed as $J_{ij} \sim \mathcal{N}\left(0,\frac{1}{Np}\right)$ where $Np$ is the average degree. We generated 5 random systems where the average degree was $Np = 100(0.02) = 2$. Across all of the systems, we used Swendsen-Wang sampling \cite{swendsen1987nonuniversal} to sample synthetic data and checked that the sampling was sufficient to eliminate autocorrelation in the data. 

For inference, we tested both $L_1$-regularized deterministic approaches as well as a variational approach based on Persistent VI. The $L_1$ regularized approaches included Pseudolikelihood, (PL) \cite{aurell2012inverse}, Minimum Probability Flow (MPF) \cite{sohl2011new}, and Persistent Contrastive Divergence (PCD) \cite{tieleman2008training}. Additionally, we tested the proposed alternative regularization method of Pseudolikelihood Decimation \cite{decelle2014pseudolikelihood}. 

For $L_1$ regularized Pseudolikelihood and Minimum Probability Flow, we selected the hyperparameter $\lambda$ using 10-fold cross-validation over 10 logarithmically spaced values on the interval $[0.01, 10]$. We performed $L_1$ regularization of the deterministic objectives using optimizers from \cite{schmidt2010graphical}, and chose the corresponding $L_1$ hyperparameter for PCD + $L_1$ based on the optimal cross-validated value of $\lambda$ that was selected for $L_1$-regularized Pseudolikelihood.

For the hierarchical model inferred with Persisent VI, we placed a separate noncentered Horseshoe prior over the fields and couplings, in accodance with the (centered) hierarchy
\begin{align*}
s_h &  \sim \operatorname{C}^+(0,1), & s_J &  \sim \operatorname{C}^+(0,1), \\
\sigma_i &  \sim \operatorname{C}^+(0,s_h), & \sigma_{ij} &\sim \operatorname{C}^+(0,s_J),\\
h_i &  \sim \mathcal{N}(0, \sigma_i^2),  & J_{ij} &\sim \mathcal{N}(0, \sigma_{ij}^2).
\end{align*}
where $\operatorname{C}^+(0,1)$ is the standard Half-Cauchy distribution. We then used PVI-3 with 100 persistent Markov chains and performed stochastic gradient descent using Adam \cite{kingma2014adam} with default momentum and a learning rate that linearly decayed from 0.01 to 0 over $5\times 10^4$ iterations. 

\subsection{Synthetic Protein Data}
We constructed a synthetic Potts spin glass with sparse interactions chosen to reflect an underlying 3D structure. After forming a contact topology from a random packed polymer, we generated synthetic group-Student-t distributed sitewise bias vectors ${\bf h}_i$ (each $20 \times 1$) and Gaussian distributed coupling matrices ${\bf J}_{ij}$ (each $20 \times 20$) to mirror the strong site-bias and weak-coupling regime of proteins. Since this system is highly frustrated, we thinned $2\times 10^6$ sweeps of Gibbs sampling to 2000 sequences that exhibited no autocorrelation.

Given 400 of the 2000 synthetic sequences\footnote{We find this effective sample size to mirror natural protein families (unpublished)}, we inferred $L_2$ and group $L_1$-regularized MAP estimates under a pseudolikelihood approximation with 5-fold cross validation to choose hyperparameters from 6 values in the range $\{0.3, 1.0, 3.0, 10.0, 30.0, 100.0\}$. We also ran PVI-10 with 40 persistent Markov chains and 5000 iterations of stochastic gradient descent with Adam\footnote{$\alpha = 0.01, \beta_1 = 0.9, \beta_2 = 0.999$, no decay} \cite{kingma2014adam}. We note that the current standards of the field are based on $L_2$ and Group $L_1$ regularized Pseudolikelihood \cite{balakrishnan2011learning, ekeberg2013improved}.

\subsection{Real Protein Data}

\subsubsection{Sample reweighting}
Natural protein sequences share a common evolutionary history that introduces significant redundancy and correlation between related sequences. Treating them as independent data is biased by both the overrepresentation of certain sequences due to the evolutionary process (phylogeny) or the human sampling process (biased sequencing of particular species). Thus, we follow a standard practice of correcting the overrepresentation of sequences by a sample-reweighting approach \cite{ekeberg2013improved}.

\paragraph{Sequence reweighting.} If we were to treat all data as independent, then every sample would receive unit weight in the log likelihood. To correct for the over and underrepresentation of certain sequences, we estimate relative sequence weights using a common inverse neighborhood density based approach from the field \cite{ekeberg2013improved}. We set the relative weight of each sequence proportional to the inverse number of neighboring sequences that differ by a normalized Hamming distance of less than $\theta$. We use the established value of $\theta = 0.2$. 
 
\paragraph{Effective sample size estimation.} We propose a new definition for an effective sample size $N_{eff}$ of correlated discrete data and derive an algorithm for estimating it from count data. The estimator is based on the assumption that in limited data regimes for sparsely coupled systems, the sample Mutual Information between random variables is dominated by random, coincidental correlations rather than actual correlations due to underlying interactions. This is consistent with classic results on the bias of information quantities in limited data regimes known as ``Miller Maddow bias" \cite{miller1955note,paninski2003estimation}. If we can define a null model for how such coincidental correlations would arise for a given random sample of size $N$, then we define $N_{eff}$ as the sample size that matches the expected null MI to the observed MI.

\begin{equation}
\mathbb{E}_{i,j}\left[\textrm{MI}_{null}| N_{eff}\right] = \mathbb{E}_{i,j}\left[\textrm{MI}_{data}\right]
\end{equation}

The expectation on the right is given by the average sample Mutual Information in the data, while the expectation on the left will be specific to a null model for Mutual Information $\mathbb{E}_{i,j}\left[MI_{null}| N\right]$. Given a noisy estimator for $\mathbb{E}_{i,j}\left[MI_{null}| N_{eff}\right]$, we solve for $N_{eff}$ by matching the expectations with Robbins-Monro stochastic optimization. 

To define the null model of mutual information $\mathbb{E}_{i,j}\left[MI_{null}| N\right]$ we treat every variable as independent categorical counts that were drawn from a Dirichlet-Multinomial hierarchy with a log-uniform hyperprior over the (symmetric) concentration parameter $\alpha$. 

Given observed frequencies ${\bf f}_i$ and ${\bf f}_j$ for letters $x_i$ and $x_j$ together with a candidate sample size $N$, we (i) use Bayes' theorem to sample underlying distributions ${\bf p}_i$, ${\bf p}_j$ that produced the observed frequencies, (ii) generate $N$ samples from the null joint distribution  ${\bf p}_i{\bf p}_j^T$, and (iii) compute the sample Mutual Information of this synthetic count data (Algorithm \ref{sampleMI}). 

We also experimented with using both MAP and posterior mean estimators as plugin approximations ${\bf \hat p}_i$, ${\bf \hat p}_j$ for the latent distributions, but found that each of these were biased estimators of the true sample size in simulation. Posterior mode estimates generally underestimated the null entropy (${\bf \hat p}_i$ too rough) while the posterior mean overestimated the entropy (${\bf \hat p}_i$ too smooth). It seems reasonable that this would be the behavior of point estimates that do not account for the uncertainty in the null distributions that is signaled by the roughness of the frequency data.

We note that this estimator will become invalid as the data become strong, since the assumption that Mutual Information is dominated by sampling noise will break down. However, for the real protein data that we examined, we found that this approach for effective sample size correction was critical for Bayesian methods such as Peristent VI to be able to set the top level hyperparameters (the sparsity levels) from the data.

\begin{algorithm}[htbp]
   \caption{Sample the null mutual information as a function of sample size  $\mathbb{E}_{i,j}\left[MI_{null}| N\right]$}
   \label{sampleMI}
   \begin{algorithmic}
   \STATE {\bfseries Require:} Sample size $N$
   \STATE {\bfseries Require:} Observed frequencies ${\bf f}_{i}$, ${\bf f}_{j}$
   \STATE Sample positions $i \in [L]$, $j \in [L]\setminus i$
   \STATE Set count data ${\bf C}_{i} \gets N {\bf f}_{i}$, ${\bf C}_{j} \gets N {\bf f}_{j}$
   \STATE Sample concentration parameter $\alpha_i | {\bf C}_{i}$, $\alpha_j | {\bf C}_{j}$ with numerical CDF
   \STATE Sample null distributions ${\bf p}_i | {\bf C}_{i}, \alpha_i$, ${\bf p}_j | {\bf C}_{j}, \alpha_j$ from Dirichlet
   \STATE Sample joint count data ${\bf M}(x_i,x_j)$ from categorical joint distribution ${\bf p}_i{\bf p}_j^T$
   \STATE Compute sample frequencies ${\bf f} = \frac{1}{N} {\bf M}(x_i,x_j)$, ${\bf f}_i = \frac{1}{N} \sum_{x_j}{\bf M}(x_i,x_j)$, ${\bf f}_j = \frac{1}{N} \sum_{x_i}{\bf M}(x_i,x_j)$
   \STATE Compute sample Mutual Information $MI = \sum_{x_i,x_j} {\bf f}(x_i,x_j) \log\frac{{\bf f}(x_i,x_j)}{{\bf f}_i(x_i){\bf f}_j(x_j)}$
\end{algorithmic}
\end{algorithm}

\subsubsection{Inference and results}
\paragraph{Alignment} Our sequence alignment was based on the Pfam 27.0 family PF00018, which we subsequently processed to remove all sequences with more than 25\% gaps.

\paragraph{Indels} Natural sequences contain insertions and deletions that are coded by `gaps' in alignments. We treated these as a 21st character (in addition to amino acids) and fit a $q = 21$ state Potts model. We acknowledge that, while this may be standard practice in the field, it is a strong independence approximation because all of the gaps in deletions are perfectly correlated.

\paragraph{Inference} We used 10,000 iterations of PVI-10 with 10 variational samples per iteration and 40 persistent Gibbs chains.

\paragraph{Comparison to 3D structure} We collected about $260$ 3D structures of SH3 domains referenced on PF00018 (Pfam 27.0) and computed minimum atom distances between all positions in the Pfam alignment. For each pair $i,j$, we used the median of distances across all structures to summarize the ``typical" minimum atom distance between $i$ and $j$.

\end{document}